\documentclass[10pt,twocolumn,letterpaper]{article}

\usepackage[pagenumbers]{cvpr} 

\usepackage{times}
\usepackage{epsfig}
\usepackage{graphicx}
\usepackage{amsmath}
\usepackage{amssymb}

\usepackage{multirow}
\usepackage{enumitem}
\usepackage{booktabs} 
\usepackage{colortbl}
\usepackage{xcolor}
\usepackage{pifont} 
\newcommand{\cmark}{\ding{51}}%
\newcommand{\xmark}{\ding{55}}%

\usepackage[pagebackref=true,breaklinks=true,colorlinks,bookmarks=false]{hyperref}

\def\paperID{695}
\def\confName{CVPR}
\def\confYear{2024}

\begin{document}

\title{PELA: Learning Parameter-Efficient Models with Low-Rank Approximation}

\author{
Yangyang Guo,
Guangzhi Wang,
Mohan Kankanhalli \\
National University of Singapore
}

\maketitle

\begin{abstract}
Applying a pre-trained large model to downstream tasks is prohibitive under resource-constrained conditions.
Recent dominant approaches for addressing efficiency issues involve adding a few learnable parameters to the fixed backbone model.
This strategy, however, leads to more challenges in loading large models for downstream fine-tuning with limited resources.
In this paper, we propose a novel method for increasing the parameter efficiency of pre-trained models by introducing an intermediate pre-training stage.
To this end, we first employ low-rank approximation to compress the original large model and then devise a feature distillation module and a weight perturbation regularization module.
These modules are specifically designed to enhance the low-rank model.
In particular, we update only the low-rank model while freezing the backbone parameters during pre-training. 
This allows for direct and efficient utilization of the low-rank model for downstream fine-tuning tasks.
The proposed method achieves both efficiencies in terms of required parameters and computation time while maintaining comparable results with minimal modifications to the backbone architecture.
Specifically, when applied to three vision-only and one vision-language Transformer models, our approach often demonstrates a merely $\sim$0.6 point decrease in performance while reducing the original parameter size by 1/3 to 2/3.
The code has been released at \href{https://github.com/guoyang9/PELA}{link}.
\end{abstract}

\section{Introduction}
Pre-training a large model and fine-tuning it at hand has become a \emph{de facto} paradigm in diverse research fields~\cite{bert, vit, ofa}.
While significant performance has been achieved, building such models often compromises increased memory usage and longer training time.
Despite these challenges, recent advances in the appreciation of the scaling law~\cite{scaling-law} and emergent abilities~\cite{emergent} of language pre-training have further fueled practitioners' interest in developing and utilizing large models.

As it is usually prohibitive to deploy these models for downstream tasks, recent studies have resorted to bypassing the fine-tuning of the entire model.
Typical approaches often introduce a few more learnable parameters to the backbone model while freezing the rest, \eg, adapter~\cite{adapter} and prompt tuning~\cite{prompt} add tunable parameters to the middle and peripheral token positions of Transformers, respectively.
However, this approach inevitably leads to the following two disadvantages.
First, the potential of pre-trained large models is not fully exploited as the majority of parameters are not tuned with downstream task objectives.
Second, loading the pre-trained model becomes even more burdensome for researchers with limited resources.
In contrast, conventional methods such as knowledge distillation (KD)~\cite{kd-relu, kd, pkd} and quantization~\cite{quan1, quan2} can partially alleviate this issue.
Yet, there is currently no established approach for constructing a high-performing student model of KD and non-differentiable operators of quantization usually make it less feasible to perform back-propagation.

\begin{figure}[t!]
  \centering
  \includegraphics[width=1.0\linewidth]{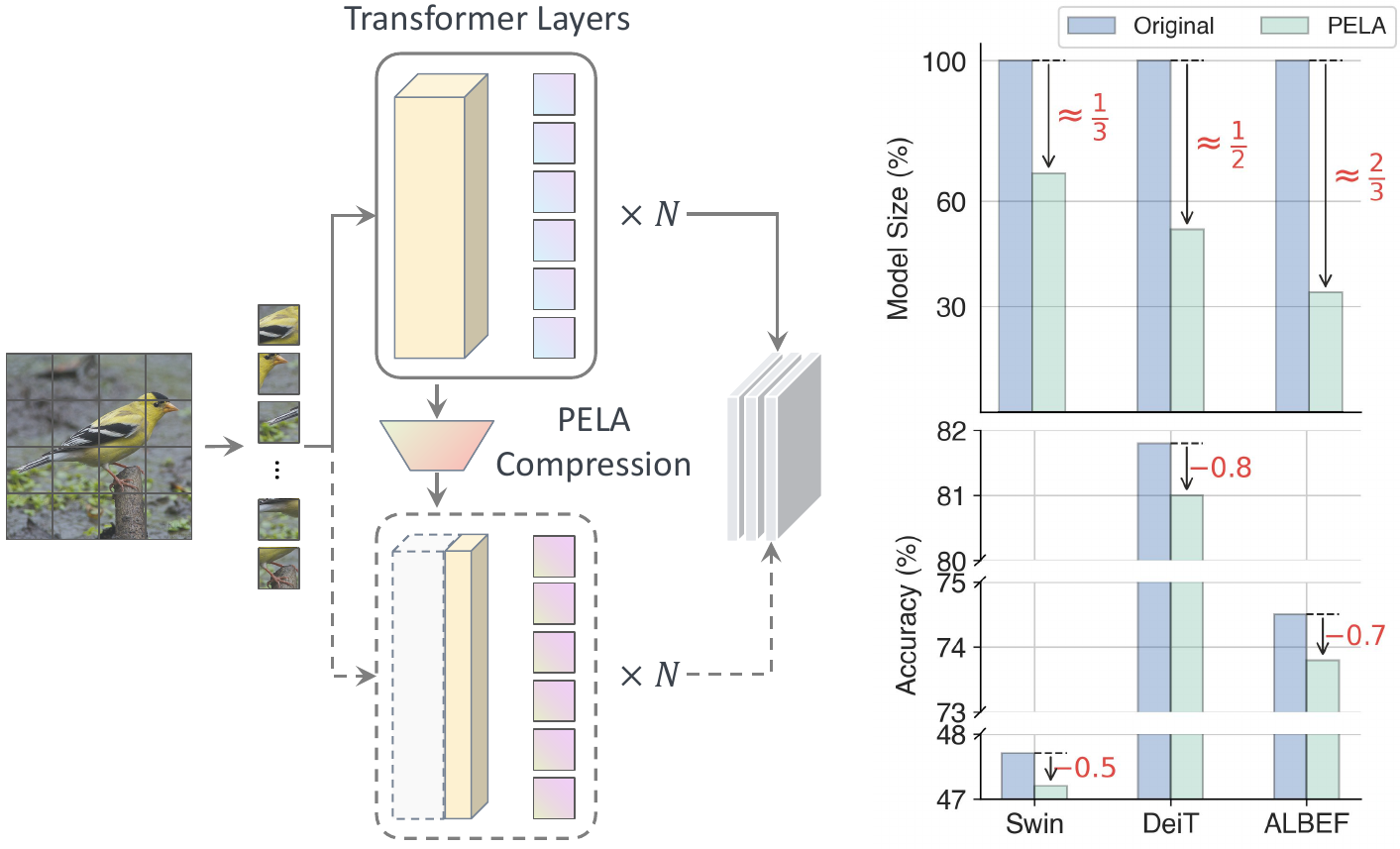}
  \vspace{-1em}
  \caption{Overview and performance of our proposed PELA method.
  Left: Using PELA, we compress the trainable weights of a typical ViT model while preserving its overall architecture.
  Right: Comparison of the Original and our PELA \emph{w.r.t} relative model size and accuracy metric on three pre-trained Transformers.}\label{fig:teaser}
  \vspace{-1em}
\end{figure}

This paper targets developing a highly parameter-efficient approach to help downstream task fine-tuning, as illustrated in Fig.~\ref{fig:teaser}.
By parameter-efficiency, we refer to a compressed model with reduced-size (\eg, 2$\times$ smaller), easy-to-implement, computationally-efficient, and minimal-architectural-change merits.
Our method offers a pre-trained compressed model that downstream tasks can directly perform fine-tuning on, in contrast to previous approaches such as LoRA~\cite{lora}, adapter~\cite{adapter}, and prompt tuning~\cite{prompt}.
In particular, this method is specially designed to address the over-parameterization problem~\cite{over-param}, where we resort to the low-rank approximation to replace the pre-trained weight matrix in each matrix multiplication operation with two low-rank matrices.
By this means, the original model size and fine-tuning time are both fairly reduced.
However, using this naive approach to perform fine-tuning yields less satisfactory outcomes (refer to Sec.~\ref{sec:exp_ablation}).
We attribute this fact to two reasons: 
Directly decomposition with low-rank approximation cannot effectively learn instance-level discriminative representations;
and the intermediate feature distribution is perturbed after this operation, resulting in sub-optimal performance.

In order to approach this problem, we propose fully taking advantage of the pre-trained model via two modules.
The implementation involves two parallel model branches: one consists of the pre-trained model with fixed parameters during pre-training, while the other is the low-rank model with tunable parameters.
Based upon this framework, our first module distills the feature knowledge from the large pre-trained model to our compressed low-rank model in terms of each Transformer layer.
The other module helps bind the weight change within a pre-defined perturbation radius.
These two modules help the low-rank model mimic the feature distribution of the large pre-trained model, thereby enhancing its discrimination capability.
During fine-tuning, we simply use the low-rank model as a replacement for the original large one to achieve parameter and computational efficiency for downstream tasks.

As far as we know, the literature on achieving a desirable efficiency-effectiveness trade-off by using low-rank approximation on pre-trained Transformer weights is quite limited. 
We apply our method to three vision-only Transformers \ie, DeiT~\cite{deit}, DeiT-III-Large~\cite{deit3} and SwinT~\cite{swin}, with 1/2 to 2/3 of the original model parameters; and one vision-language Transformer -- ALBEF~\cite{albef} where the parameter size is reduced to 1/3 of the original model.
We then conduct extensive experiments on a range of downstream tasks, including image classification, semantic segmentation, and object detection for the vision-only Transformers;
Visual entailment, visual grounding, cross-modal retrieval, and visual question answering for the vision-language Transformer.
Our approach achieves performance that is highly comparable to the backbone model, with differences mostly around 0.6 points, despite using only 1/3 to 2/3 of the original FLOPs.
In addition, this parameter-efficiency benefit further enables the model to scale with larger batch sizes, leading to improved performance that sometimes even outperforms backbones.

\section{Related Work}
\subsection{Parameter-Efficient Learning}
Efficiency has long been an engaging problem in a variety of research areas~\cite{pre-dl1, pre-dl2}.
After stepping into the deep representation learning era, the progressive improvements in our community often trade with a large number of model parameters, latency, and footprints~\cite{efficient-survey, prune-survey}.
With this concern, previous efforts have been mainly devoted to three distinctive directions: knowledge distillation (KD), quantization, and pruning.
Deemed as a principled model compression algorithm, early KD aims to transfer the knowledge from a cumbersome teacher model to a lightweight student model via class logit alignment~\cite{kd, distill-bert}.
Recent focus has been shifted to feature-based knowledge transfer due to its performance advantage over conventional logit-based ones~\cite{kd-relu, tiny-bert, fitnet, kd-abandon}.
For example,~\cite{tiny-bert, pkd} distill the knowledge from hidden states and attention matrices, which on the other hand, can also bypass the logit-free training objectives.
However, choosing features from which layers to align remains challenging as there is no teacher-student layer match from a theoretical basis.
Quantization, from another angle of efficient learning, maps larger bit parameters to smaller ones, \eg, 32-bit floating point to an 8-bit integer~\cite{quantization-survey}.
This kind of method is not dependent on the model structures, which makes it flexible in various neural networks~\cite{quan1, quan2, quan3}.
The key downside lies in its performance reduction and possible infeasibility for back-propagation.
Different from the above two categories, pruning is leveraged to remove unnecessary or less important components in models~\cite{prune-survey}. 
By removing some connections~\cite{prune-unstructured} or parameters~\cite{prune-structured}, the original dense network reduces to a sparse one, in which the required capacity for storage as well as the amount of computations will dwindle.

Transformer-based approaches have succeeded in diverse research domains since their introduction~\cite{bert, transformer}.
These models often involve billions of parameters, which consequently, motivates some specific methods working on addressing the parameter-efficiency problem~\cite{al-bert}.
The typical strategy is to add a few learnable parameters while freezing the majority of the Transformer backbone during downstream training.
For instance, prompt tuning appends some task-specific parameters into the input space~\cite{prompt};
Adapter models introduce several learnable MLP components into each Transformer layer~\cite{adapter};
and fine-tuning bias only has also been proven effective for maintaining good performance of large language models~\cite{bias-only}.

\subsection{Low-Rank Approximation}
Low-rank approximation aims to decompose one matrix into two smaller matrices, subject to the constraint that the resulting matrices have reduced rank~\cite{low-rank-old1, low-rank-old2}.
One key merit of this algorithm is data compression, whereby previous work has applied it to principal component analysis~\cite{pca} and recommendation~\cite{lr-recommendation2, lr-recommendation1}.

Pertaining to Convolutional Neural Networks (CNNs), some approaches apply the low-rank approximation to each feature map via higher-order tensor decomposition~\cite{lr-cnn1, lr-cnn2, lr-cnn3}.
Dynamically decomposing trainable matrices has also attracted much attention~\cite{lr-dynamic2, lr-dynamic1, lr-dynamic3}.
Some more studies explored other aspects of low-rank approximation, such as rank learning~\cite{rank-learn}, constrained optimization~\cite{lr-constrain}, and employing it specifically in token embedding matirx~\cite{lr-embedding, al-bert} or self-attention computation in Transformers~\cite{lin-former}.
LoRA~\cite{lora} models the residual of parameters with low-rank approximation, wherein only the newly decomposed matrices are exploited for downstream training and it thus achieves significantly reduced trainable parameters.
Despite its benefits, the LoRA approach still has limitations, as it necessitates the storage and reloading of large pre-trained weights in hard disk and GPU memory, respectively. 
In other words, only the newly introduced trainable parameters that are of a smaller magnitude compared to the full parameters are updated for fine-tuning, making it similar to adapters~\cite{adapter-enhanced, adapter} and prompt tuning~\cite{prompt}.
Unlike existing approaches, our method uses low-rank approximation during pre-training to entirely replace the pre-trained weights with reduced low-rank matrices. 
As a result, we achieve both memory and computational efficiency goals for downstream fine-tuning tasks.

\section{Method}
Transformers have grown into a fundamental building block of many modern vision models~\cite{vit, maskAE}. 
Take the seminal Vision Transformer (ViT) as an example.
ViT first divides an RGB image $I \in \mathbb{R}^{3 \times H \times W}$ into $M \times M$ non-overlapping patches.
Together with a class token, these image patches are thereafter fed into $N$ layers with self-attention as the basic operation.
To this end, a set of query, key, and value matrices are transformed from the patch embedding to token features $\mathbf{X} \in \mathbb{R}^{(M^2 + 1) \times d}$, where $d$ denotes the embedding size, followed by several feedforward layers and residual connections.
At their core lies the fully connected layer, which is often wrapped in the attention score estimation and MLP operations - $\mathbf{W}^T \mathbf{X} + \mathbf{b}$, 
where $\mathbf{W} \in \mathbb{R}^{d_{in} \times d_{out}}$ is the learnable weight matrix and $\mathbf{b} \in \mathbb{R}^{d_{out}}$ denotes the bias, and $d_{in} = d$ for the first layer.


\begin{figure}[t!]
  \centering
  \includegraphics[width=1.0\linewidth]{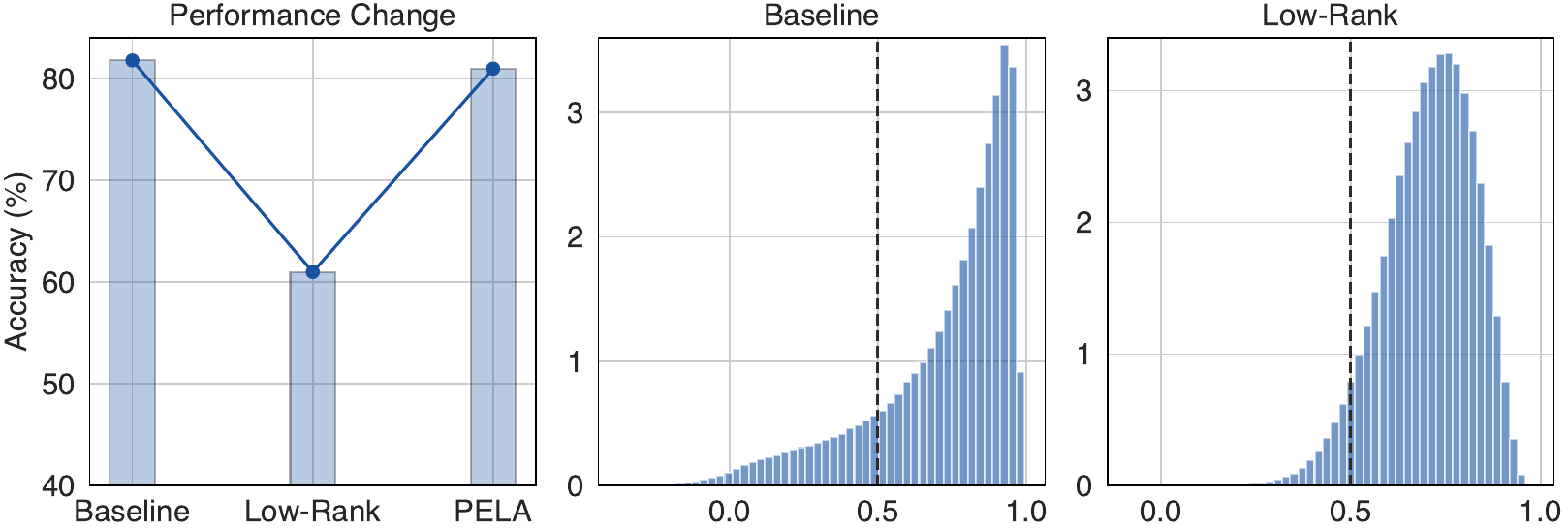}
  \vspace{-1em}
  \caption{Performance comparison of three models and statistics of the instance-level feature similarity.
  Left: We use the DeiT model as the baseline and show the performance of its directly low-rank approximation and PELA variants.
  The middle and right sub-figures illustrate the instance-level feature similarity of DeiT and directly low-rank model variants, respectively.}\label{fig:hist}
  \vspace{-1em}
\end{figure}

\begin{figure*}[t!]
  \centering
  \includegraphics[width=0.9\linewidth]{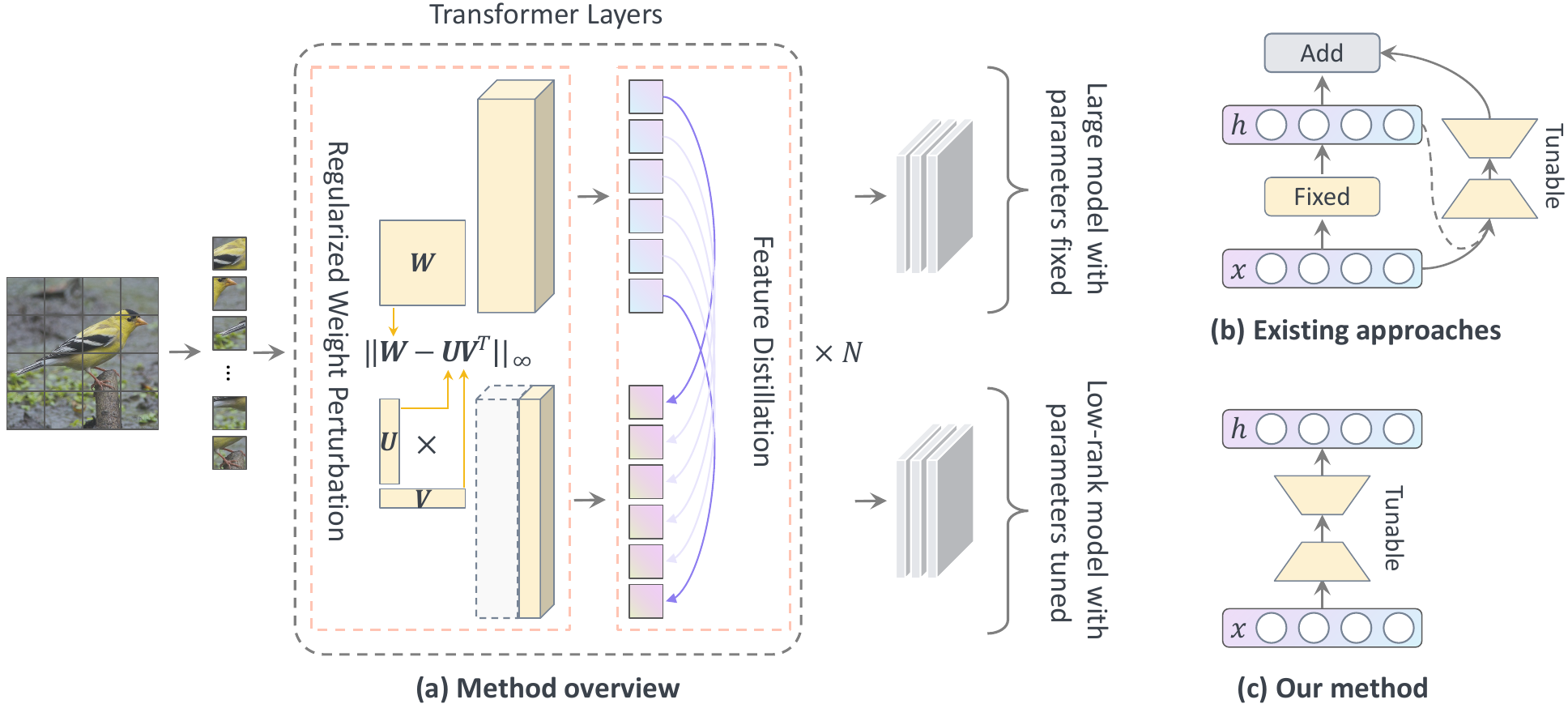}
  \caption{Overview of our proposed PELA and pipeline comparison with existing methods.
  (a) We leverage a typical ViT model as the base for the illustration of our method.
  The two involved modules, \ie, Feature Distillation aligns the token features in an apple-to-apple fashion of each layer, and Regularized Weight Perturbation bounds the recovered weight matrices.
  During fine-tuning on downstream tasks, existing approaches use both the fixed pre-trained model weights and the newly added parameters (b). 
  In contrast, our PELA keeps only the low-rank model while excluding the large pre-trained model for efficient computation (c).
  }\label{fig:model}
  \vspace{-1em}
\end{figure*}

\subsection{Low-rank Approximation}
Over-parameterization is a common issue in modern large models~\cite{over-param}.
In this work, we aim to address this problem by reducing the number of model parameters. 
Inspired by the success of low-rank approximation in other domains~\cite{lr-cnn1, lr-cnn2}, we propose to apply this technique directly to the matrix multiplication operations in ViT,
\begin{equation} \label{eqn:lr}
\begin{split}
    \mathbf{W}^T \mathbf{X} & \approx (\mathbf{U} \mathbf{V}^T)^T \mathbf{X} \\
                            & = \mathbf{V} (\mathbf{U}^T \mathbf{X}), \\
\end{split}
\end{equation}
where $\mathbf{U} \in \mathbb{R}^{d_{in} \times d_{lr}}$ and $\mathbf{V} \in \mathbb{R}^{d_{out} \times d_{lr}}$ are low-rank matrices, and $d_{lr}$ represents the desired rank of $\mathbf{W}$.
Note that the weight matrices in a deep learning model are often with full-rank, \ie, $rank(\mathbf{W}) = min(d_{in}, d_{out})$.
Under such conditions, we seek approximately equal the original matrix and deliberately choose a smaller $d_{lr}$, \eg, $\frac{1}{4} min(d_{in}, d_{out})$.
The second equation constantly holds in neural networks due to the natural associative law.
This property allows us to achieve computational efficiency without needing to recover the original weight matrix $\mathbf{W}$ after applying the low-rank approximation.
We utilize the well-known SVD approach~\cite{svd} to perform the low-rank approximation as,
\begin{equation}
    \mathtt{SVD}(\mathbf{W}^T) = \mathbf{U}^* \mathbf{\Sigma} \mathbf{V}^*,  
\end{equation}
where $\mathbf{\Sigma} \in \mathbb{R}^{d_{in} \times d_{out}}$ is a rectangular diagonal matrix with non-negative real numbers on the diagonal, and the singular values are sorted in a monotonously decreasing order; $\mathbf{U^*} \in \mathbb{R}^{d_{in} \times d_{in}}$ and $\mathbf{V^*} \in \mathbb{R}^{d_{out} \times d_{out}}$ are complex unitary matrices.
We then formalize the low-rank matrices using the following transformation,
\begin{equation}
    \begin{cases}
    \mathbf{U} = \mathbf{U}^*_{[:, : d_{lr}]} \mathbf{\Sigma}^{\frac{1}{2}}_{[:d_{lr}, : d_{lr}]}, \\
    \mathbf{V} = (\mathbf{\Sigma}^{\frac{1}{2}}_{[:d_{lr}, : d_{lr}]} \mathbf{V}^*_{[:d_{lr}, :]})^T, \\
    \end{cases}
\end{equation}
where $[:, :d_{lr}]$ implies we truncate the given matrix with the top-$d_{lr}$ columns and other truncation operations can also be easily deduced.

\noindent \textbf{Preliminary observation.} 
We apply this low-rank approximation to the fully connected layers of pre-trained models.
Unfortunately, this process delivers less desirable results, \eg, the accuracy drops from 81\% to 61\% as seen in Fig.~\ref{fig:hist}.
This indicates that the compressed low-rank model does not effectively learn instance-level discriminative representation.
Moreover, we found that the learned features after low rank are confined in a narrow feature space.
In particular, the right two subfigures in Fig.~\ref{fig:hist} demonstrate that the feature similarity of each class of the low-rank model is drastically higher than before.

To overcome this, we propose to take full advantage of the large pre-trained model and harness it to guide the training of the low-rank model.
Specifically, as shown in Fig.~\ref{fig:model}, we first perform low-rank approximation on the pre-trained model and retain both models.
The parameters of the large pre-trained model are frozen while we train only the low-rank model.
Our method further involves two modules: \emph{feature distillation} to align features between these two models, \emph{regularized weight perturbation} to constrain the affinity of the recovered matrix and original matrix. 
We name this method \textbf{PELA}, dubbed \emph{\textbf{P}arameter-\textbf{E}fficient models for \textbf{L}ow-rank \textbf{A}pproximation}.
To the best of our knowledge, there is limited research on constructing an effective low-rank model based on pre-trained Transformers. 
Therefore, we aim to address this gap by investigating the potential of low-rank approximation to achieve an optimal efficiency-effectiveness trade-off. 

\subsection{Feature Distillation}
As highlighted in the previous sub-section, the low-rank approximation can alter the feature distribution of the pre-trained model.
To address this problem, we resort to feature-based knowledge distillation, which has been proven effective in aligning the features between models~\cite{fitnet}.
Nevertheless, the low-rank compression is performed on each matrix multiplication operation, rather than specific Transformer blocks or layers.
Directly distilling knowledge from all the output features of the original model leads to more clutter as some low-rank compression is already wrapped within the self-attention computation.
Thanks to the layer-wise residual connection of Transformers, we employ a compromise in this work -- simply aligning the token features of each layer.
From a general view of typical Transformer models, the feature distillation loss is defined as follows,
\begin{equation}
\begin{split}
    \mathcal{L}_{fd}    &= \sum_{i=1}^N \mathcal{D} (\mathcal{M}_s (\mathbf{X}^i_s), \mathcal{M}_t (\mathbf{X}^i_t)), \\
                        &= \frac{1}{2N} \sum_{i=1}^N \parallel \mathcal{M}_s (\mathbf{X}^i_s) - \mathcal{M}_t (\mathbf{X}^i_t) \parallel^2,
\end{split}
\end{equation}
where $\mathbf{X}^i_s$ and $\mathbf{X}^i_t$ denote the $i$-th layer token features of the compressed low-rank model and original model, respectively; $\mathcal{M}$ is a transformation that transfers the feature to the target feature space and we employ identity mapping in our implementation.
In this way, the output features from each ViT layer of the low-rank model are expected to share a similar distribution with that of the corresponding large pre-trained model.

\noindent \textbf{An alternative view from knowledge distillation.}
Recent studies have shown that feature-based knowledge distillation significantly outperforms conventional logit-based one~\cite{kd-relu}.
Nonetheless, how to design a student model and transfer the knowledge from the teacher model remains challenging as it is rather difficult to define teacher-student feature matching.
Our method offers a neat solution to this problem due to the following two reasons:
1) Unlike previous approaches (\eg,~\cite{tiny-bert, pkd} manually removing certain layers), the low-rank approximation is effortless and straightforward to compress the cumbersome teacher model to a lightweight student model.
2) There exists a natural correspondence between the teacher model and student model as we have not excessively altered the model architectures.  

\subsection{Regularized Weight Perturbation}
An ideal low-rank approximation is to learn an approximating matrix of the original one subject to a reduced rank constraint.
This leads to an efficiency-effectiveness dilemma -- A larger rank corresponds to a lower reconstruction error and vice versa.
Intuitively, we associate the matrix reconstruction with that of weight perturbation, which is relatively new as opposed to the feature/input perturbation robustness problem~\cite{weight-perturb}.
As a result, a smaller rank in our method, from the other angle, can be seen as more weight perturbations.
To reduce the negative influence of these perturbed parameters, we use the $l_\infty$-norm for constraining the reconstruction error,
\begin{equation}
\begin{cases}
    \parallel \hat{\mathbf{W}}^{(k)} - \mathbf{W}^{(k)} \parallel_\infty \leq \epsilon, \\
    \hat{\mathbf{W}}^{(k)} = \mathbf{U}^{(k)} (\mathbf{V}^{(k)})^T, \forall k \in [{K}]
\end{cases}    
\end{equation}
where $\epsilon$ represents the perturbation radius, $\mathbf{W}^{(k)}$ is the original weight matrix, and $[{K}]$ denotes the weight index set.
Given $\epsilon$, preserving the neural network robustness against weight perturbation can be cast as the following optimization problem~\cite{weight-perturb},
\begin{equation}
    \mathcal{L}_{rwp} = \sum_{k=1}^{\mid [K] \mid} (\parallel \hat{\mathbf{W}}^{(k)} - \mathbf{W}^{(k)} \parallel_\infty - \epsilon).
\end{equation}

\subsection{Training}
The above two modules enable us to capture the compelling discriminative capability of large pre-trained models.
To obtain a compact low-rank model, we comprehensively consider the objectives from both the base pre-training and our proposed two modules,  
\begin{equation}
    \mathcal{L} = \mathcal{L}_{base} + \alpha \mathcal{L}_{fd} + \beta \mathcal{L}_{rwp},
\end{equation}
where $\alpha$ and $\beta$ are loss weight hyper-parameters and $\mathcal{L}_{base}$ is the loss functions of the original pre-training tasks.
It can be the classification loss of a typical ViT, or vision-text matching and masked language modeling losses of a vision-language model.
We then optimize our model on the same datasets as the pre-trained model, such as ImageNet~\cite{imagenet}.

After this intermediate pre-training stage, our low-rank model is smoothly deployed for downstream fine-tuning since the model architecture is rarely altered.
In contrast to existing methods such as prompt tuning~\cite{prompt}, adapters~\cite{adapter}, and LoRA~\cite{lora}, which require both the large pre-trained model and the fine-tuning parameters, we only keep the low-rank model for efficient inference and parameter usage (see Fig.~\ref{fig:model} for a visual comparison).

\subsection{Complexity Analysis}
Before analyzing the complexity of our method, we first let $d_{lr}=\frac{1}{\kappa} \frac{d_{in} \times d_{out}}{d_{in} + d_{out}}$, where $\kappa$ is a positive number and we name it \emph{compression ratio}.
We choose to use a universal compression ratio for all the matrix multiplication operations for simplicity while leaving the exploration of dynamic ratios for different layers as future work.

Let us consider the case where $\kappa=2$ and a single patch feature $x \in \mathbb{R}^{d_{in}}$ for downstream fine-tuning.
Recall Eqn.~\ref{eqn:lr}, the original matrix multiplication takes $\mathcal{O}(d_{in} \times d_{out})$ to operate.
However, with our PELA method, this time complexity reduces to $\mathcal{O}((d_{in} + d_{out}) \times d_{lr}) = \frac{1}{2}\mathcal{O}(d_{in} \times d_{out})$\footnote{We refer this reduced complexity only to the matrix multiplication since we do not optimize other operations such as attention computation.}.
Similarly, as the majority operation in existing Transformers is matrix multiplication (excluding some very few layer normalization parameters and bias parameters), the model size thus also roughly halves from its original scale.
This is why our method is significantly different from other recent efficient approaches such as LoRA~\cite{lora}, where the overall model size in fact increases.

\section{Experiments}
\subsection{Common Efficient Learning Baselines}
We evaluated our PELA against four efficient baselines: \textbf{TinyBERT~\cite{tiny-bert}} and \textbf{MaskAlign~\cite{maskalign}} from the feature-based knowledge distillation group;
\textbf{ToMe~\cite{tome}} - a recent strong vision token pruning approach; and \textbf{LoRA~\cite{lora}}, which is a widely used parameter-efficient transfer learning baseline.
However, we excluded some experiments due to certain incompatibilities, such as using ToMe for the Swin model and for the visual grounding task.

\begin{table}[htbp]
    \centering
    \caption{Model performance of image classification on ImageNet-1K~\cite{imagenet} with 224x224 resolution.
    The parameters and FLOPs are estimated during inference.}
    \resizebox{0.9\linewidth}{!}{
    \begin{tabular}{c|c|c|c|c}
    \toprule
    \multicolumn{2}{c|}{Method}                             & Params(M)                    & GFLOPs                         & Acc(\%)                   \\
    \midrule
    \multicolumn{2}{c|}{ViT-Base~\cite{vit}}                & 86.6                          & 35.1                          & 77.9                      \\
    \multicolumn{2}{c|}{CrossViT-B~\cite{crossvit}}         & 105.0                         & 40.3                          & 82.2                      \\
    \multicolumn{2}{c|}{T2T-ViT-24~\cite{t2t-vit}}          & 64.1                          & 25.5                          & 82.3                      \\
    \multicolumn{2}{c|}{RegNetY-16G~\cite{regnet}}          & 83.6                          & 31.9                          & 82.9                      \\
    \midrule
    \multirow{5}{*}{DeiT}
        & Base~\cite{deit}                                  & 86.6                          & 33.7                          & 81.8                      \\
        \cmidrule(lr){2-5}
        & \cellcolor{pink!15}TinyBert~\cite{tiny-bert}      & \cellcolor{pink!15}44.2       & \cellcolor{pink!15}17.3       & \cellcolor{pink!15}78.0   \\
        & \cellcolor{pink!15}MaskAlign~\cite{maskalign}     & \cellcolor{pink!15}44.2       & \cellcolor{pink!15}17.3       & \cellcolor{pink!15}78.2   \\
        & \cellcolor{pink!15}ToMe~\cite{tome}               & \cellcolor{pink!15}86.6       & \cellcolor{pink!15}16.5       & \cellcolor{pink!15}76.4   \\
        & \cellcolor{gray!15}PELA                           & \cellcolor{gray!15}44.1       & \cellcolor{gray!15}17.0       & \cellcolor{gray!15}81.0   \\
    \midrule
    \multirow{4}{*}{Swin-Base}               
        & Base~\cite{swin}                                  & 87.8                          & 30.3                          & 83.5                      \\
        \cmidrule(lr){2-5}
        & \cellcolor{pink!15}TinyBert~\cite{tiny-bert}      & \cellcolor{pink!15}58.6       & \cellcolor{pink!15}20.6       & \cellcolor{pink!15}78.8   \\
        & \cellcolor{pink!15}MaskAlign~\cite{maskalign}     & \cellcolor{pink!15}58.6       & \cellcolor{pink!15}20.6       & \cellcolor{pink!15}79.1   \\
        & \cellcolor{gray!15}PELA                           & \cellcolor{gray!15}62.2       & \cellcolor{gray!15}21.3       & \cellcolor{gray!15}82.5   \\
    \midrule
    \multirow{4}{*}{DeiT-III-Large}        
        & Base~\cite{deit3}                                 & 304.4                         & 119.4                         & 84.9                      \\
        \cmidrule(lr){2-5}
        & \cellcolor{pink!15}TinyBert~\cite{tiny-bert}      & \cellcolor{pink!15}156.8      & \cellcolor{pink!15}61.5       & \cellcolor{pink!15}79.2   \\
        & \cellcolor{pink!15}MaskAlign~\cite{maskalign}     & \cellcolor{pink!15}156.8      & \cellcolor{pink!15}61.5       & \cellcolor{pink!15}79.5   \\
        & \cellcolor{gray!15}PELA                           & \cellcolor{gray!15}153.2      & \cellcolor{gray!15}59.8       & \cellcolor{gray!15}83.9   \\    
    \bottomrule
    \end{tabular}} 
    \label{tab:classification}
\end{table}

\begin{table}[htbp]
    \centering
    \caption{Model performance of semantic segmentation on the ADE20K dataset~\cite{ade20k} with UperNet~\cite{upernet}.}
    \resizebox{0.9\linewidth}{!}{
    \begin{tabular}{c|c|c|c|c}
    \toprule
    \multicolumn{2}{c|}{Backbone}                           & Params(M)                 & GFLOPs                        & mIoU                      \\
    \midrule
    \multicolumn{2}{c|}{ResNet-101~\cite{resnet}}           & 85.5                      & 689                           & 44.9                      \\
    \multicolumn{2}{c|}{PatchConvNet-B60~\cite{patchconv}}  & \textcolor{gray}{141.0}   & \textcolor{gray}{1,258}       & 48.1                      \\
    \multicolumn{2}{c|}{MAE ViT-B~\cite{maskAE}}            & 163.9                     & 2,343                         & 48.1                      \\
    \midrule
    \multirow{5}{*}{DeiT}
        & Base~\cite{deit}                                  & 121.4                     & 320.4                         & 45.0                      \\
        & \textcolor{gray}{LoRA~\cite{lora}}                & \textcolor{gray}{124.8}   & \textcolor{gray}{331.1}       & \textcolor{gray}{40.6}    \\
        \cmidrule(lr){2-5}
        & \cellcolor{pink!15}TinyBert~\cite{tiny-bert}      & \cellcolor{pink!15}79.0   & \cellcolor{pink!15}214.5      & \cellcolor{pink!15}36.4       \\
        & \cellcolor{pink!15}MaskAlign~\cite{maskalign}     & \cellcolor{pink!15}79.0   & \cellcolor{pink!15}214.5      & \cellcolor{pink!15}36.8       \\
        & \cellcolor{gray!15}PELA                           & \cellcolor{gray!15}78.9   & \cellcolor{gray!15}203.4      & \cellcolor{gray!15}43.2   \\
    \midrule
    \multirow{5}{*}{Swin-Base}
        & Base~\cite{swin}                                  & 121.3                     & 798.6                         & 47.7                      \\
        & \textcolor{gray}{LoRA~\cite{lora}}                & \textcolor{gray}{124.7}   & \textcolor{gray}{822.6}       & \textcolor{gray}{44.2}    \\
        \cmidrule(lr){2-5}
        & \cellcolor{pink!15}TinyBert~\cite{tiny-bert}      & \cellcolor{pink!15}92.1   & \cellcolor{pink!15}721.4      & \cellcolor{pink!15}40.0       \\
        & \cellcolor{pink!15}MaskAlign~\cite{maskalign}     & \cellcolor{pink!15}92.1   & \cellcolor{pink!15}721.4      & \cellcolor{pink!15}39.6       \\
        & \cellcolor{gray!15}PELA                           & \cellcolor{gray!15}79.3   & \cellcolor{gray!15}685.3      & \cellcolor{gray!15}47.2   \\
    \midrule
   \multirow{5}{*}{DeiT-III-Large}
        & Base~\cite{deit3}                                 & 428.4                     & 1,155                         & \textcolor{gray}{47.0}    \\
        & \textcolor{gray}{LoRA~\cite{lora}}                & \textcolor{gray}{440.4}   & \textcolor{gray}{1,190}       & \textcolor{gray}{44.7}    \\
        \cmidrule(lr){2-5}
        & \cellcolor{pink!15}TinyBert~\cite{tiny-bert}      & \cellcolor{pink!15}280.8  & \cellcolor{pink!15}784        & \cellcolor{pink!15}38.1       \\
        & \cellcolor{pink!15}MaskAlign~\cite{maskalign}     & \cellcolor{pink!15}280.8  & \cellcolor{pink!15}784        & \cellcolor{pink!15}38.4       \\
        & \cellcolor{gray!15}PELA                           & \cellcolor{gray!15}277.2  & \cellcolor{gray!15}739        & \cellcolor{gray!15}45.6   \\
    \bottomrule
    \end{tabular}} 
    \label{tab:segmentation}
    \vspace{-1em}
\end{table}

\subsection{Experiments on Vision-Only Models}
\subsubsection{Baseline Models and Results}
We applied our method to the widely used DeiT-Base~\cite{deit} and Swin-Base~\cite{swin} models.
To ensure comprehensive coverage, we also selected DeiT-III-Large~\cite{deit3} which is larger in model size and requires much longer training time.
The compression ratio is 1/2 and 1/3 for the DeiT models and Swin, respectively.
After the low-rank approximation, we trained our model on the ImageNet-1k dataset~\cite{imagenet} and evaluated it on the corresponding validation set, and report the results in Table~\ref{tab:classification}.
As expected, the model parameters and FLOPs \emph{for inference} are significantly reduced according to each respective compression ratio.
On the flip side, the dropped accuracy of the two base models is 0.8\% and 1.0\%, respectively.
Even for the relatively larger model DeiT-III-Large, our method only trades 1.0\% accuracy with half of the parameters and FLOPs.
Moreover, our PELA surpasses other efficient learning baselines by a notable margin.

\begin{table}[htbp]
    \centering
    \caption{Model performance of object detection on the MSCOCO dataset~\cite{coco} with Cascade Mask RCNN~\cite{cascade-rcnn, mask-rcnn}.}
    \resizebox{0.9\linewidth}{!}{
    \begin{tabular}{c|c|c|c|c}
    \toprule
    \multicolumn{2}{c|}{Backbone}                       & Params(M)                 & GFLOPs                    & $\text{AP}^{box}$         \\
    \midrule
    \multicolumn{2}{c|}{ResNet-50~\cite{resnet}}        & 77.3                      & 411.0                     & 46.3                      \\  
    \multicolumn{2}{c|}{ResNeXt-101-32~\cite{resnext}}  & 96.0                      & 546.1                     & 48.1                      \\
    \midrule
    \multirow{5}{*}{Swin-Base}
        & Base~\cite{swin}                              & 145.0                     & 1,501                     & 50.1                      \\
        & \textcolor{gray}{LoRA~\cite{lora}}            & \textcolor{gray}{149.0}   & \textcolor{gray}{1,547}   & \textcolor{gray}{46.1}    \\
        \cmidrule(lr){2-5}
        & \cellcolor{pink!15}TinyBert~\cite{tiny-bert}  & \cellcolor{pink!15}115.8       & \cellcolor{pink!15}1,302       & \cellcolor{pink!15}41.1       \\
        & \cellcolor{pink!15}MaskAlign~\cite{maskalign} & \cellcolor{pink!15}115.8       & \cellcolor{pink!15}1,302       & \cellcolor{pink!15}41.1       \\
        & \cellcolor{gray!15}PELA                       & \cellcolor{gray!15}103.0  & \cellcolor{gray!15}1,232  & \cellcolor{gray!15}49.0   \\
    \bottomrule
    \end{tabular}} 
    \label{tab:detection}
    \vspace{-1em}
\end{table}

\subsubsection{Downstream Tasks and Results}
After the backbones are pre-trained on the ImageNet dataset, as per prior studies~\cite{maskAE, resnet}, we further evaluated the model performance on downstream semantic segmentation and object detection tasks.

\begin{table*}[htbp]
    \centering
    \caption{Performance comparison of text retrieval (TR) and image Retrieval (IR) on the Flickr30K and MSCOCO datasets.} 
    \resizebox{0.8\linewidth}{!}{
    \begin{tabular}{c|c|c|c|c|ccc|ccc}
    \toprule
    \multirow{2}{*}{Dataset}    & \multicolumn{2}{c|}{\multirow{2}{*}{Model}}       & \multirow{2}{*}{Params}   & \multirow{2}{*}{TFLOPs}   & \multicolumn{3}{c|}{TR}   & \multicolumn{3}{c}{IR}    \\
                                                                                                                \cmidrule(lr){6-8}          \cmidrule(lr){9-11}
                                                       
                                & \multicolumn{2}{c|}{}                             &                           &                           & R@1   & R@5   & R@10      & R@1   & R@5   & R@10      \\
    \midrule
    \multirow{8}{*}{Flickr30K}  & \multicolumn{2}{c|}{UNITER~\cite{uniter}}         & 110                       & 0.37                      & 87.3  & 98.0  & 99.2      & 75.6  & 94.1  & 96.8      \\
                                & \multicolumn{2}{c|}{VILLA~\cite{villa}}           & 110                       & -                         & 87.9  & 97.5  & 98.8      & 76.3  & 94.2  & 96.8      \\
                                \cmidrule(lr){2-11}
                                & \multirow{6}{*}{ALBEF}
                                & Base~\cite{albef}                                 & 419                       & 7.41                      & 93.4  & 99.5  & 99.6      & 80.6  & 95.8  & 98.0     \\
                                && \textcolor{gray}{LoRA~\cite{lora}}               & \textcolor{gray}{431}     & \textcolor{gray}{7.49}  
                                                                                    & \textcolor{gray}{92.1}    & \textcolor{gray}{99.2}    & \textcolor{gray}{99.0}
                                                                                    & \textcolor{gray}{80.2}    & \textcolor{gray}{95.6}    & \textcolor{gray}{97.7}                                \\
                                \cmidrule(lr){3-11}
                                && \cellcolor{pink!15}TinyBERT~\cite{tiny-bert}     & \cellcolor{pink!15}230    & \cellcolor{pink!15}4.66                      
                                                                                    & \cellcolor{pink!15}57.6   & \cellcolor{pink!15}82.8   & \cellcolor{pink!15}89.9      
                                                                                    & \cellcolor{pink!15}40.8   & \cellcolor{pink!15}70.6   & \cellcolor{pink!15}79.4                               \\
                                && \cellcolor{pink!15}MaskAlign~\cite{maskalign}    & \cellcolor{pink!15}230    & \cellcolor{pink!15}4.66                         
                                                                                    & \cellcolor{pink!15}59.0   & \cellcolor{pink!15}84.2   & \cellcolor{pink!15}90.9      
                                                                                    & \cellcolor{pink!15}41.1   & \cellcolor{pink!15}70.4   & \cellcolor{pink!15}80.5                               \\
                                && \cellcolor{pink!15}ToMe~\cite{tome}              & \cellcolor{pink!15}419    & \cellcolor{pink!15}2.61
                                                                                    & \cellcolor{pink!15}74.8   & \cellcolor{pink!15}92.6   & \cellcolor{pink!15}96.4      
                                                                                    & \cellcolor{pink!15}62.0   & \cellcolor{pink!15}86.2   & \cellcolor{pink!15}91.5                               \\
                                && \cellcolor{gray!15}PELA                          & \cellcolor{gray!15}173    & \cellcolor{gray!15}2.58     
                                                                                    & \cellcolor{gray!15}91.6   & \cellcolor{gray!15}99.3   & \cellcolor{gray!15}99.6
                                                                                    & \cellcolor{gray!15}79.7   & \cellcolor{gray!15}94.8   & \cellcolor{gray!15}97.5                               \\
    \midrule
    \multirow{8}{*}{MSCOCO}     & \multicolumn{2}{c|}{UNITER~\cite{uniter}}         & 110                       & 0.37                      & 65.7  & 88.6  & 93.8      & 52.9  & 79.9  & 88.0      \\
                                & \multicolumn{2}{c|}{OSCAR~\cite{oscar}}           & 110                       & -                         & 70.0  & 91.1  & 95.5      & 54.0  & 80.8  & 88.5      \\
                                \cmidrule(lr){2-11}
                                & \multirow{6}{*}{ALBEF}
                                & Base~\cite{albef}                                 & 419                       & 7.41                      & 72.6  & 91.2  & 95.2      & 54.9  & 80.5  & 88.1     \\
                                && \textcolor{gray}{LoRA~\cite{lora}}               & \textcolor{gray}{431}     & \textcolor{gray}{7.49}  
                                                                                    & \textcolor{gray}{73.2}    & \textcolor{gray}{91.7}    & \textcolor{gray}{95.9}
                                                                                    & \textcolor{gray}{56.5}    & \textcolor{gray}{81.3}    & \textcolor{gray}{88.9}                                \\
                                \cmidrule(lr){3-11}
                                && \cellcolor{pink!15}TinyBERT~\cite{tiny-bert}     & \cellcolor{pink!15}230    & \cellcolor{pink!15}4.66                          
                                                                                    & \cellcolor{pink!15}33.6   & \cellcolor{pink!15}62.1   & \cellcolor{pink!15}74.8      
                                                                                    & \cellcolor{pink!15}22.6   & \cellcolor{pink!15}49.8   & \cellcolor{pink!15}63.2                               \\
                                && \cellcolor{pink!15}MaskAlign~\cite{maskalign}    & \cellcolor{pink!15}230    & \cellcolor{pink!15}4.66                         
                                                                                    & \cellcolor{pink!15}35.7   & \cellcolor{pink!15}64.9   & \cellcolor{pink!15}77.3      
                                                                                    & \cellcolor{pink!15}24.2   & \cellcolor{pink!15}52.5   & \cellcolor{pink!15}65.5                               \\
                                && \cellcolor{pink!15}ToMe~\cite{tome}              & \cellcolor{pink!15}419    & \cellcolor{pink!15}2.61
                                                                                    & \cellcolor{pink!15}56.2   & \cellcolor{pink!15}82.3   & \cellcolor{pink!15}90.1     
                                                                                    & \cellcolor{pink!15}41.7   & \cellcolor{pink!15}71.0   & \cellcolor{pink!15}81.3                               \\
                                && \cellcolor{gray!15}PELA                          & \cellcolor{gray!15}173    & \cellcolor{gray!15}2.58     
                                                                                    & \cellcolor{gray!15}71.6   & \cellcolor{gray!15}91.0   & \cellcolor{gray!15}95.3
                                                                                    & \cellcolor{gray!15}55.1   & \cellcolor{gray!15}80.8   & \cellcolor{gray!15}88.3                               \\
    \bottomrule
    \end{tabular}}
    \vspace{-1em}
    \label{tab:retrieval}
\end{table*}

The results are presented in Table~\ref{tab:segmentation} and Table~\ref{tab:detection}, which illustrate the effectiveness of our method in performing semantic segmentation and object detection tasks, respectively. 
While our approach benefits from the reduced memory and computation requirements, the involvement of downstream frameworks and heads limits the extent to which these benefits can be realized when compared to vanilla classification. 
For instance, the reduced FLOPs for Swin-Base on object detection in Table~\ref{tab:detection} are 18\% as compared to the previous 30\% in Table~\ref{tab:classification}.
Nevertheless, our approach still performs comparably with each respective model, demonstrating its effectiveness in balancing the trade-off between efficiency and accuracy. 
Notably, our PELA significantly outperforms LoRA in terms of both model performance and model size.

\begin{table}[htbp]
    \centering
    \caption{Model performance on visual entailment and VQA.
    Params (M) and TFLOPs are counted based on the VQA model.} 
    \resizebox{1.0\linewidth}{!}{
    \begin{tabular}{c|c|c|c|cc|cc}
    \toprule
    \multicolumn{2}{c|}{\multirow{2}{*}{Model}}         & \multirow{2}{*}{Params}   & \multirow{2}{*}{TFLOPs}   & \multicolumn{2}{c|}{SNLI-VE}  & \multicolumn{2}{c}{VQA}   \\
                                                                                                                \cmidrule(lr){5-6}              \cmidrule(lr){7-8}
    \multicolumn{2}{c|}{}                               &                           &                           &  val          & test          & test-dev  & test-std      \\
    \midrule
    \multicolumn{2}{c|}{VisualBERT~\cite{visualbert}}   & 134                       & 0.37                      & -             & -             & 70.80     & 71.00         \\
    \multicolumn{2}{c|}{ViLT~\cite{vilt}}               & 118                       & 1.01                      & -             & -             & 70.94     & -             \\
    \multicolumn{2}{c|}{LXMERT~\cite{lxmert}}           & 224                       & 0.41                      & -             & -             & 72.42     & 72.54         \\
    \multicolumn{2}{c|}{UNITER~\cite{uniter}}           & 116                       & 0.37                      & 78.59         & 78.28         & 72.70     & 72.91         \\
    \multicolumn{2}{c|}{12-in-1~\cite{12in1}}           & -                         & -                         & -             & 76.59         & 73.15     & -             \\
    \midrule
    \multirow{5}{*}{ALBEF}     
        & Base                                          & 581                       & 7.05                      & 79.29         & 79.79         & 74.55     & 74.89         \\
        &  \textcolor{gray}{LoRA}                       & \textcolor{gray}{644}     & \textcolor{gray}{7.14}                      
                                                        & \textcolor{gray}{79.34}   & \textcolor{gray}{79.53}   & \textcolor{gray}{71.07}       & \textcolor{gray}{-}       \\
        \cmidrule(lr){2-8}
        & \cellcolor{pink!15}TinyBERT                   & \cellcolor{pink!15}392    & \cellcolor{pink!15}4.55                      
                                                        & \cellcolor{pink!15}73.83  & \cellcolor{pink!15}73.31  & \cellcolor{pink!15}61.33      & \cellcolor{pink!15}-      \\
        & \cellcolor{pink!15}MaskAlign                  & \cellcolor{pink!15}392    & \cellcolor{pink!15}4.55                     
                                                        & \cellcolor{pink!15}73.74  & \cellcolor{pink!15}73.48  & \cellcolor{pink!15}63.85      & \cellcolor{pink!15}-      \\
        & \cellcolor{pink!15} ToMe                      & \cellcolor{pink!15}581    & \cellcolor{pink!15}2.55
                                                        & \cellcolor{pink!15}77.58  & \cellcolor{pink!15}78.02  & \cellcolor{pink!15}68.59      & \cellcolor{pink!15} -     \\
        & \cellcolor{gray!15}PELA                       & \cellcolor{gray!15}259    & \cellcolor{gray!15}2.47     
                                                        & \cellcolor{gray!15}78.55  & \cellcolor{gray!15}78.66  & \cellcolor{gray!15}73.84      & \cellcolor{gray!15}73.87  \\
    \bottomrule
    \end{tabular}}
    \label{tab:vqa}
\end{table}

\subsection{Experiments on Vision-Language Model}
\subsubsection{Baseline Model and Downstream VL Tasks}
Traditional visual-language pre-training approaches~\cite{uniter, lxmert} frequently utilized pre-extracted CNN features for image representation, often requiring precise bounding box annotations.
In contrast, ALBEF~\cite{albef} leverages ViT for visual feature extraction during pre-training and has exhibited exceptional performance across a variety of VL tasks.
Therefore, we chose ALBEF as our evaluation testbed to assess the effectiveness of our method.
Furthermore, the all-in Transformer nature of ALBEF enabled us to effortlessly achieve more compression.
In this context, we used 1/3 of the parameters of the original ALBEF model.

We utilized four downstream vision-language tasks in this work, including Image-Text Retrieval, SNLI-VE, VG, and VQA. 
A detailed introduction to these tasks can be found in the supplementary material.
For the experiments, we strictly followed the implementation of ALBEF except for reducing the batch size due to resource constraints.

\subsubsection{Overall Results}
The results on these downstream tasks are reported in Table~\ref{tab:retrieval},~\ref{tab:vqa}, and~\ref{tab:grounding}.
From these tables, we have the following three important observations.
1) The recent approach ALBEF~\cite{albef} has demonstrated significant performance improvements over conventional methods like LXMERT~\cite{lxmert} and UNITER~\cite{uniter}.
However, superior performance is achieved at the expense of increased parameters and FLOPs, mainly due to the usage of a cumbersome trainable ViT for image processing.
In comparison to the baselines, which use a universal Transformer for both vision and language, such as UNITER~\cite{uniter}, ALBEF offers superior visual features but introduces a larger model size and computational complexity.
2) Our PELA method helps alleviate this problem through the low-rank approximation.
As can be observed, PELA is able to achieve comparable performance to ALBEF while using only 1/3 of the parameters and FLOPs. 
This translates to a significant reduction in model size and computation, with most performance degradation limited to just one point.
\begin{table}[htbp]
    \centering
    \caption{Model performance on the challenging weakly-supervised visual grounding task.} 
    \resizebox{0.6\linewidth}{!}{
    \begin{tabular}{c|c|ccc}
    \toprule
    \multicolumn{2}{c|}{Model}          & Val                       & TestA                     & TestB                     \\
    \midrule
    \multicolumn{2}{c|}{ARN~\cite{arn}} & 32.78                     & 34.35                     & 32.13                     \\
    \multicolumn{2}{c|}{CCL~\cite{ccl}} & 34.29                     & 36.91                     & 33.56                     \\
    \midrule
    \multirow{2}{*}{ALBEF}     
        & Base                          & 57.94                     & 65.07                     & 45.75                     \\
        \cmidrule(lr){2-5}
        & \cellcolor{gray!15}PELA       & \cellcolor{gray!15}57.06  & \cellcolor{gray!15}65.85  & \cellcolor{gray!15}45.10 \\
    \bottomrule
    \end{tabular}}
    \vspace{-1em}
    \label{tab:grounding}
\end{table}
3) Regarding the comparison with efficient learning baselines, our PELA approach consistently achieves better performance in most cases. 
The only exception is for retrieval tasks, where PELA exhibits slightly inferior model performance compared to LoRA. 
However, it is important to note that LoRA requires a larger number of model parameters and FLOPs.

\begin{table*}[htbp]
    \centering
    \caption{Ablation studies of the proposed method over five tasks.
    For the downstream tasks of ALBEF, we selected representative evaluation metrics for space concerns.} 
    \resizebox{0.78\linewidth}{!}{
    \begin{tabular}{c|cc|cc|cc|cc|cc|cc}
    \toprule
    \multirow{3}{*}{Model}  & \multirow{3}{*}{$\mathcal{L}_{fd}$}   & \multirow{3}{*}{$\mathcal{L}_{rwp}$}        
                                                    & \multicolumn{2}{c|}{DeiT} & \multicolumn{2}{c|}{Swin} & \multicolumn{6}{c}{ALBEF}                                                                     \\
                                                    \cmidrule(lr){4-5}          \cmidrule(lr){6-7}          \cmidrule(lr){8-13}
                                                    &&& Cls         & Seg       & Cls       & Seg           & \multicolumn{2}{c|}{Retrieval}    & \multicolumn{2}{c|}{SNLI-VE}  & \multicolumn{2}{c}{VG}    \\
                                                    \cmidrule(lr){4-5}          \cmidrule(lr){6-7}          \cmidrule(lr){8-9}                  \cmidrule(lr){10-11}            \cmidrule(lr){12-13}
                                                    &&& Acc         & mIoU      & Acc       & mIoU          & TR@1     & IR@1                   & val       & test              & TestA     & TestB         \\
    \midrule
    \multicolumn{3}{c|}{\textcolor{gray}{Baseline}} & \textcolor{gray}{81.80}   & \textcolor{gray}{44.99}   & \textcolor{gray}{83.50}   & \textcolor{gray}{47.68}   & \textcolor{gray}{72.64}           
                                                    & \textcolor{gray}{54.91}   & \textcolor{gray}{79.29}   & \textcolor{gray}{79.79}   & \textcolor{gray}{65.07}   & \textcolor{gray}{45.75}               \\
    \midrule
    \multirow{4}{*}{PELA}   & \xmark    & \xmark    & 61.08         & 24.42     & 77.60     & 28.80         & 65.94     & 49.45                 & 76.10     & 76.21             & 61.36     & 41.36         \\
                            & \cmark    & \xmark    & 80.90         & 43.67     & 82.89     & 47.28         & 70.66     & 54.55                 & 78.35     & 78.07             & 65.70     & 44.90         \\
                            & \xmark    & \cmark    & 80.55         & 42.94     & 82.86     & 47.24         & 71.44     & 54.43                 & 78.50     & 78.39             & 66.24     & 44.57         \\  
                            & \cmark    & \cmark    & 80.96         & 43.24     & 82.54     & 47.21         & 71.26     & 54.75                 & 78.55     & 78.66             & 65.86     & 45.10         \\                                                 
    \bottomrule
    \end{tabular}
    }
    \label{tab:ablation}
\end{table*}

\begin{figure*}[t!]
  \centering
  \includegraphics[width=1.0\linewidth]{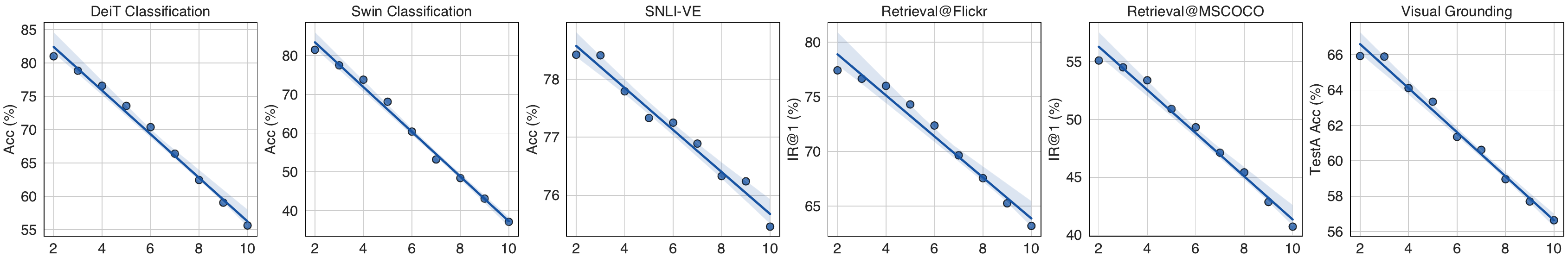}
  \caption{Model performance change \emph{w.r.t.} compression ratios.}\label{fig:ratio}
  \vspace{-1em}
\end{figure*}

\begin{table}[htbp]
    \centering
    \caption{Efficiency comparison of two pre-training strategies.} 
    \resizebox{0.8\linewidth}{!}{
    \begin{tabular}{c|c|cc}
    \toprule
        Method                          & Batch Size                    & GPU Mem$\downarrow$       & Latency$\downarrow$   \\ 
    \midrule
        DeiT-Base                       & \multirow{2}{*}{64$\times$4}  & 12.77 GB                  & 76.24 ms/img          \\
        $\textbf{DeiT-Base}_{pela}$     &                               & 11.34 GB                  & 70.06 ms/img         \\
    \bottomrule
    \end{tabular}}
    \label{tab:efficiency}
\end{table}

\subsection{Ablation Study} \label{sec:exp_ablation}
\noindent \textbf{Effectiveness of the two modules.} 
We first studied the model performance of direct decomposition of pre-trained weights using low-rank approximation.
However, as indicated in Table~\ref{tab:ablation}, this approach results in a significant drop in performance, possibly because of the shift in feature distribution. 
We then added our proposed two modules to the low-rank model and observed performance improvements.
By combining the two modules together, our model can often outperform other variants, demonstrating the effectiveness of the proposed method.

\noindent \textbf{Performance variation \emph{w.r.t.} compression ratio.} 
Training large models often involves a trade-off between effectiveness and efficiency. 
To demonstrate this, we trained our model using different compression ratios with fewer epochs to simplify the process and present the results in Fig.~\ref{fig:ratio}. 
This graph indicates that a smaller compression ratio, \ie, a larger model, typically yields better performance. 
However, a model that is too small, such as one that is compressed to 1/10 of its original size, may not be capable of achieving satisfactory results.

\subsection{Pre-training Efficiency \& Model Scaling}
\noindent \textbf{Pre-training Efficiency.}
One may be concerned about the efficiency issues during pre-training.
To address this problem, we leveraged the DeiT-Base model and evaluated its pre-training efficiency metrics, and show the results in Table~\ref{tab:efficiency}.
In particular, we employed the plain low-rank model because it already delivers promising model performance.
Though other models may trigger longer training time, under this context, as shown in the table, our PELA method outperforms the original model in terms of both GPU memory cost and training latency.

\noindent \textbf{Downstream Model Scaling.}
Our approach spawns a more compact model compared to the original large pre-trained one, resulting in a surplus of memory that enables us to train downstream models with larger batch sizes. 
To demonstrate the effectiveness of our method, we increased the batch size for both DeiT-Base and Swin-Base on the semantic segmentation task\footnote{We kept the GPU memory less than the original large baseline model for a fair comparison.}, as shown in Table~\ref{tab:scale}. 
\begin{table}[htbp]
    \centering
    \caption{Model scaling performance of DeiT-Base and Swin-Base on semantic segmentation.
    PELA+ denotes the PELA model with a larger batch size while maintaining similar GPU memory.} 
    \resizebox{0.8\linewidth}{!}{
    \begin{tabular}{c|c|ccc}
    \toprule
        Method              & Batch Size            & Baseline                          & PELA      & PELA+                             \\ 
    \midrule
        DeiT-Base           & 16$\rightarrow$20     & \textcolor{gray}{44.99}           & 43.24     & 43.81$_{\textcolor{blue}{+.57}}$    \\
        Swin-Base           & 16$\rightarrow$20     & \textcolor{gray}{47.68}           & 47.21     & 47.99$_{\textcolor{blue}{+.78}}$  \\
    \bottomrule
    \end{tabular}}
    \vspace{-1em}
    \label{tab:scale}
\end{table}
Our experiments show promising results, with a significant improvement in model performance for both models, achieving an absolute mIoU improvement of 0.57\% and 0.78\%, respectively. 
Moreover, our proposed method also outperforms the original Swin-Base baseline using PELA+, highlighting another advantage of our proposed approach.

\section{Conclusion and Future Work}
In this work, we propose a simple yet effective parameter-efficient pre-training approach that employs low-rank approximation as the core.
Even with its simplicity, our method achieves competitive performance with baselines while attaining significantly improved parameter and computational efficiencies.
These advantages enable model scaling in terms of model depth, width, and training batch size of downstream task fine-tuning.
This work highlights the potential benefits of tackling the over-parameterization problem of learnable weights.
In addition to this, we believe that the compression of intermediate features is a promising orthogonal direction for reducing model complexity. 
Therefore, we plan to investigate feature compression techniques, such as vision token pruning, to further build a more lightweight model in future research.

{\small
\bibliographystyle{ieee_fullname}
\bibliography{pela}

\begin{thebibliography}{75}
\providecommand{\natexlab}[1]{#1}
\providecommand{\url}[1]{\texttt{#1}}
\expandafter\ifx\csname urlstyle\endcsname\relax
  \providecommand{\doi}[1]{doi: #1}\else
  \providecommand{\doi}{doi: \begingroup \urlstyle{rm}\Url}\fi

\bibitem[Balaji et~al.(2021)Balaji, Sajedi, Kalibhat, Ding, St{\"{o}}ger,
  Soltanolkotabi, and Feizi]{over-param}
Yogesh Balaji, Mohammadmahdi Sajedi, Neha~Mukund Kalibhat, Mucong Ding, Dominik
  St{\"{o}}ger, Mahdi Soltanolkotabi, and Soheil Feizi.
\newblock Understanding overparameterization in generative adversarial
  networks.
\newblock In \emph{ICLR}, 2021.

\bibitem[Bolya et~al.(2023)Bolya, Fu, Dai, Zhang, Feichtenhofer, and
  Hoffman]{tome}
Daniel Bolya, Cheng{-}Yang Fu, Xiaoliang Dai, Peizhao Zhang, Christoph
  Feichtenhofer, and Judy Hoffman.
\newblock Token merging: Your vit but faster.
\newblock In \emph{ICLR}, 2023.

\bibitem[Cai and Vasconcelos(2018)]{cascade-rcnn}
Zhaowei Cai and Nuno Vasconcelos.
\newblock Cascade r-cnn: Delving into high quality object detection.
\newblock In \emph{CVPR}, pages 6154--6162. IEEE, 2018.

\bibitem[Chen et~al.(2021)Chen, Fan, and Panda]{crossvit}
Chun-Fu~Richard Chen, Quanfu Fan, and Rameswar Panda.
\newblock Crossvit: Cross-attention multi-scale vision transformer for image
  classification.
\newblock In \emph{ICCV}, pages 357--366. IEEE, 2021.

\bibitem[Chen et~al.(2018)Chen, Si, Li, Chelba, and Hsieh]{lr-embedding}
Patrick~H. Chen, Si Si, Yang Li, Ciprian Chelba, and Cho{-}Jui Hsieh.
\newblock Groupreduce: Block-wise low-rank approximation for neural language
  model shrinking.
\newblock In \emph{NeurIPS}, pages 11011--11021, 2018.

\bibitem[Chen et~al.(2020)Chen, Li, Yu, Kholy, Ahmed, Gan, Cheng, and
  Liu]{uniter}
Yen{-}Chun Chen, Linjie Li, Licheng Yu, Ahmed~El Kholy, Faisal Ahmed, Zhe Gan,
  Yu Cheng, and Jingjing Liu.
\newblock {UNITER:} universal image-text representation learning.
\newblock In \emph{ECCV}, pages 104--120. Springer, 2020.

\bibitem[Courbariaux et~al.(2015)Courbariaux, Bengio, and David]{quan1}
Matthieu Courbariaux, Yoshua Bengio, and Jean{-}Pierre David.
\newblock Binaryconnect: Training deep neural networks with binary weights
  during propagations.
\newblock In \emph{NIPS}, pages 3123--3131, 2015.

\bibitem[Deng et~al.(2009)Deng, Dong, Socher, Li, Li, and Fei-Fei]{imagenet}
Jia Deng, Wei Dong, Richard Socher, Li-Jia Li, Kai Li, and Li Fei-Fei.
\newblock Imagenet: A large-scale hierarchical image database.
\newblock In \emph{CVPR}, pages 248--255. IEEE, 2009.

\bibitem[Devlin et~al.(2019)Devlin, Chang, Lee, and Toutanova]{bert}
Jacob Devlin, Ming{-}Wei Chang, Kenton Lee, and Kristina Toutanova.
\newblock {BERT:} pre-training of deep bidirectional transformers for language
  understanding.
\newblock In \emph{NAACL}, pages 4171--4186. ACL, 2019.

\bibitem[Dosovitskiy et~al.(2021)Dosovitskiy, Beyer, Kolesnikov, Weissenborn,
  Zhai, Unterthiner, Dehghani, Minderer, Heigold, Gelly, Uszkoreit, and
  Houlsby]{vit}
Alexey Dosovitskiy, Lucas Beyer, Alexander Kolesnikov, Dirk Weissenborn,
  Xiaohua Zhai, Thomas Unterthiner, Mostafa Dehghani, Matthias Minderer, Georg
  Heigold, Sylvain Gelly, Jakob Uszkoreit, and Neil Houlsby.
\newblock An image is worth 16x16 words: Transformers for image recognition at
  scale.
\newblock In \emph{ICLR}. OpenReview.net, 2021.

\bibitem[Gan et~al.(2020)Gan, Chen, Li, Zhu, Cheng, and Liu]{villa}
Zhe Gan, Yen{-}Chun Chen, Linjie Li, Chen Zhu, Yu Cheng, and Jingjing Liu.
\newblock Large-scale adversarial training for vision-and-language
  representation learning.
\newblock In \emph{NeurIPS}, 2020.

\bibitem[Guo et~al.(2018)Guo, Li, Lin, Chen, and Li]{lr-cnn1}
Jianbo Guo, Yuxi Li, Weiyao Lin, Yurong Chen, and Jianguo Li.
\newblock Network decoupling: From regular to depthwise separable convolutions.
\newblock In \emph{BMVC}, page 248. {BMVA} Press, 2018.

\bibitem[Guo et~al.(2022)Guo, Cheng, Jing, Lin, Nie, and
  Wang]{lr-recommendation2}
Yangyang Guo, Zhiyong Cheng, Jiazheng Jing, Yanpeng Lin, Liqiang Nie, and Meng
  Wang.
\newblock Enhancing factorization machines with generalized metric learning.
\newblock \emph{TKDE}, 34\penalty0 (8):\penalty0 3740--3753, 2022.

\bibitem[He et~al.(2022{\natexlab{a}})He, Zhou, Ma, Berg{-}Kirkpatrick, and
  Neubig]{adapter-enhanced}
Junxian He, Chunting Zhou, Xuezhe Ma, Taylor Berg{-}Kirkpatrick, and Graham
  Neubig.
\newblock Towards a unified view of parameter-efficient transfer learning.
\newblock In \emph{ICLR}, 2022{\natexlab{a}}.

\bibitem[He et~al.(2016)He, Zhang, Ren, and Sun]{resnet}
Kaiming He, Xiangyu Zhang, Shaoqing Ren, and Jian Sun.
\newblock Deep residual learning for image recognition.
\newblock In \emph{CVPR}, pages 770--778. IEEE, 2016.

\bibitem[He et~al.(2017{\natexlab{a}})He, Gkioxari, Doll{\'a}r, and
  Girshick]{mask-rcnn}
Kaiming He, Georgia Gkioxari, Piotr Doll{\'a}r, and Ross Girshick.
\newblock Mask r-cnn.
\newblock In \emph{ICCV}, pages 2961--2969. IEEE, 2017{\natexlab{a}}.

\bibitem[He et~al.(2022{\natexlab{b}})He, Chen, Xie, Li, Doll{\'{a}}r, and
  Girshick]{maskAE}
Kaiming He, Xinlei Chen, Saining Xie, Yanghao Li, Piotr Doll{\'{a}}r, and
  Ross~B. Girshick.
\newblock Masked autoencoders are scalable vision learners.
\newblock In \emph{CVPR}, pages 15979--15988. {IEEE}, 2022{\natexlab{b}}.

\bibitem[He et~al.(2017{\natexlab{b}})He, Liao, Zhang, Nie, Hu, and
  Chua]{lr-recommendation1}
Xiangnan He, Lizi Liao, Hanwang Zhang, Liqiang Nie, Xia Hu, and Tat{-}Seng
  Chua.
\newblock Neural collaborative filtering.
\newblock In \emph{WWW}, pages 173--182. {ACM}, 2017{\natexlab{b}}.

\bibitem[Heo et~al.(2019)Heo, Kim, Yun, Park, Kwak, and Choi]{kd-relu}
Byeongho Heo, Jeesoo Kim, Sangdoo Yun, Hyojin Park, Nojun Kwak, and Jin~Young
  Choi.
\newblock A comprehensive overhaul of feature distillation.
\newblock In \emph{ICCV}, pages 1921--1930. {IEEE}, 2019.

\bibitem[Hinton et~al.(2015)Hinton, Vinyals, and Dean]{kd}
Geoffrey~E. Hinton, Oriol Vinyals, and Jeffrey Dean.
\newblock Distilling the knowledge in a neural network.
\newblock \emph{CoRR}, abs/1503.02531, 2015.

\bibitem[Hu et~al.(2022)Hu, Shen, Wallis, Allen{-}Zhu, Li, Wang, Wang, and
  Chen]{lora}
Edward~J. Hu, Yelong Shen, Phillip Wallis, Zeyuan Allen{-}Zhu, Yuanzhi Li,
  Shean Wang, Lu Wang, and Weizhu Chen.
\newblock Lora: Low-rank adaptation of large language models.
\newblock In \emph{ICLR}. OpenReview.net, 2022.

\bibitem[Hubara et~al.(2017)Hubara, Courbariaux, Soudry, El{-}Yaniv, and
  Bengio]{quan2}
Itay Hubara, Matthieu Courbariaux, Daniel Soudry, Ran El{-}Yaniv, and Yoshua
  Bengio.
\newblock Quantized neural networks: Training neural networks with low
  precision weights and activations.
\newblock \emph{JMLR}, 18:\penalty0 187:1--187:30, 2017.

\bibitem[Idelbayev and Carreira{-}Perpi{\~{n}}{\'{a}}n(2020)]{rank-learn}
Yerlan Idelbayev and Miguel~{\'{A}}. Carreira{-}Perpi{\~{n}}{\'{a}}n.
\newblock Low-rank compression of neural nets: Learning the rank of each layer.
\newblock In \emph{CVPR}, pages 8046--8056. {IEEE}, 2020.

\bibitem[Jia et~al.(2022)Jia, Tang, Chen, Cardie, Belongie, Hariharan, and
  Lim]{prompt}
Menglin Jia, Luming Tang, Bor{-}Chun Chen, Claire Cardie, Serge~J. Belongie,
  Bharath Hariharan, and Ser{-}Nam Lim.
\newblock Visual prompt tuning.
\newblock In \emph{ECCV}, pages 709--727. Springer, 2022.

\bibitem[Jiao et~al.(2020)Jiao, Yin, Shang, Jiang, Chen, Li, Wang, and
  Liu]{tiny-bert}
Xiaoqi Jiao, Yichun Yin, Lifeng Shang, Xin Jiang, Xiao Chen, Linlin Li, Fang
  Wang, and Qun Liu.
\newblock Tinybert: Distilling {BERT} for natural language understanding.
\newblock In \emph{Findings of EMNLP}, pages 4163--4174. ACL, 2020.

\bibitem[Jr.(1998)]{pre-dl1}
Roberto J.~Bayardo Jr.
\newblock Efficiently mining long patterns from databases.
\newblock In \emph{SIGMOD}, pages 85--93. {ACM}, 1998.

\bibitem[Kaplan et~al.(2020)Kaplan, McCandlish, Henighan, Brown, Chess, Child,
  Gray, Radford, Wu, and Amodei]{scaling-law}
Jared Kaplan, Sam McCandlish, Tom Henighan, Tom~B. Brown, Benjamin Chess, Rewon
  Child, Scott Gray, Alec Radford, Jeffrey Wu, and Dario Amodei.
\newblock Scaling laws for neural language models.
\newblock \emph{CoRR}, 2020.

\bibitem[Kim et~al.(2021)Kim, Son, and Kim]{vilt}
Wonjae Kim, Bokyung Son, and Ildoo Kim.
\newblock Vilt: Vision-and-language transformer without convolution or region
  supervision.
\newblock In \emph{ICML}, pages 5583--5594. {PMLR}, 2021.

\bibitem[Kong et~al.(2022)Kong, Dong, Ma, Meng, Niu, Sun, Shen, Yuan, Ren,
  Tang, Qin, and Wang]{prune-structured}
Zhenglun Kong, Peiyan Dong, Xiaolong Ma, Xin Meng, Wei Niu, Mengshu Sun, Xuan
  Shen, Geng Yuan, Bin Ren, Hao Tang, Minghai Qin, and Yanzhi Wang.
\newblock Spvit: Enabling faster vision transformers via latency-aware soft
  token pruning.
\newblock In \emph{ECCV}, pages 620--640. Springer, 2022.

\bibitem[Lan et~al.(2020)Lan, Chen, Goodman, Gimpel, Sharma, and
  Soricut]{al-bert}
Zhenzhong Lan, Mingda Chen, Sebastian Goodman, Kevin Gimpel, Piyush Sharma, and
  Radu Soricut.
\newblock {ALBERT:} {A} lite {BERT} for self-supervised learning of language
  representations.
\newblock In \emph{ICLR}. OpenReview.net, 2020.

\bibitem[Lee et~al.(2021)Lee, Yu, Kim, Breuel, Kautz, and Song]{svd}
Sangho Lee, Youngjae Yu, Gunhee Kim, Thomas~M. Breuel, Jan Kautz, and Yale
  Song.
\newblock Parameter efficient multimodal transformers for video representation
  learning.
\newblock In \emph{ICLR}, 2021.

\bibitem[Li and Shi(2018)]{lr-constrain}
Chong Li and C.{-}J.~Richard Shi.
\newblock Constrained optimization based low-rank approximation of deep neural
  networks.
\newblock In \emph{ECCV}, pages 746--761. Springer, 2018.

\bibitem[Li et~al.(2021{\natexlab{a}})Li, Selvaraju, Gotmare, Joty, Xiong, and
  Hoi]{albef}
Junnan Li, Ramprasaath~R. Selvaraju, Akhilesh Gotmare, Shafiq~R. Joty, Caiming
  Xiong, and Steven~Chu{-}Hong Hoi.
\newblock Align before fuse: Vision and language representation learning with
  momentum distillation.
\newblock In \emph{NeurIPS}, pages 9694--9705, 2021{\natexlab{a}}.

\bibitem[Li et~al.(2019)Li, Yatskar, Yin, Hsieh, and Chang]{visualbert}
Liunian~Harold Li, Mark Yatskar, Da Yin, Cho{-}Jui Hsieh, and Kai{-}Wei Chang.
\newblock Visualbert: {A} simple and performant baseline for vision and
  language.
\newblock \emph{CoRR}, abs/1908.03557, 2019.

\bibitem[Li et~al.(2020)Li, Yin, Li, Zhang, Hu, Zhang, Wang, Hu, Dong, Wei,
  Choi, and Gao]{oscar}
Xiujun Li, Xi Yin, Chunyuan Li, Pengchuan Zhang, Xiaowei Hu, Lei Zhang, Lijuan
  Wang, Houdong Hu, Li Dong, Furu Wei, Yejin Choi, and Jianfeng Gao.
\newblock Oscar: Object-semantics aligned pre-training for vision-language
  tasks.
\newblock In \emph{ECCV}, pages 121--137. Springer, 2020.

\bibitem[Li et~al.(2021{\natexlab{b}})Li, Gong, Tan, Yang, Hu, Zhang, Yu, Wang,
  and Gu]{quan3}
Yuhang Li, Ruihao Gong, Xu Tan, Yang Yang, Peng Hu, Qi Zhang, Fengwei Yu, Wei
  Wang, and Shi Gu.
\newblock {BRECQ:} pushing the limit of post-training quantization by block
  reconstruction.
\newblock In \emph{ICLR}. OpenReview.net, 2021{\natexlab{b}}.

\bibitem[Lin et~al.(2014)Lin, Maire, Belongie, Hays, Perona, Ramanan,
  Doll{\'{a}}r, and Zitnick]{coco}
Tsung{-}Yi Lin, Michael Maire, Serge~J. Belongie, James Hays, Pietro Perona,
  Deva Ramanan, Piotr Doll{\'{a}}r, and C.~Lawrence Zitnick.
\newblock Microsoft {COCO:} common objects in context.
\newblock In \emph{ECCV}, pages 740--755. Springer, 2014.

\bibitem[Liu et~al.(2019)Liu, Li, Wang, Zha, Meng, and Huang]{arn}
Xuejing Liu, Liang Li, Shuhui Wang, Zheng{-}Jun Zha, Dechao Meng, and Qingming
  Huang.
\newblock Adaptive reconstruction network for weakly supervised referring
  expression grounding.
\newblock In \emph{ICCV}, pages 2611--2620. {IEEE}, 2019.

\bibitem[Liu et~al.(2021)Liu, Lin, Cao, Hu, Wei, Zhang, Lin, and Guo]{swin}
Ze Liu, Yutong Lin, Yue Cao, Han Hu, Yixuan Wei, Zheng Zhang, Stephen Lin, and
  Baining Guo.
\newblock Swin transformer: Hierarchical vision transformer using shifted
  windows.
\newblock In \emph{ICCV}, pages 9992--10002. {IEEE}, 2021.

\bibitem[Lu et~al.(2020)Lu, Goswami, Rohrbach, Parikh, and Lee]{12in1}
Jiasen Lu, Vedanuj Goswami, Marcus Rohrbach, Devi Parikh, and Stefan Lee.
\newblock 12-in-1: Multi-task vision and language representation learning.
\newblock In \emph{CVPR}, pages 10434--10443. {IEEE}, 2020.

\bibitem[Menghani(2021)]{efficient-survey}
Gaurav Menghani.
\newblock Efficient deep learning: {A} survey on making deep learning models
  smaller, faster, and better.
\newblock \emph{CoRR}, abs/2106.08962, 2021.

\bibitem[Papailiopoulos et~al.(2013)Papailiopoulos, Dimakis, and
  Korokythakis]{pca}
Dimitris~S. Papailiopoulos, Alexandros~G. Dimakis, and Stavros Korokythakis.
\newblock Sparse {PCA} through low-rank approximations.
\newblock In \emph{ICML}, pages 747--755. JMLR.org, 2013.

\bibitem[Radosavovic et~al.(2020)Radosavovic, Kosaraju, Girshick, He, and
  Doll{\'a}r]{regnet}
Ilija Radosavovic, Raj~Prateek Kosaraju, Ross Girshick, Kaiming He, and Piotr
  Doll{\'a}r.
\newblock Designing network design spaces.
\newblock In \emph{CVPR}, pages 10428--10436. IEEE, 2020.

\bibitem[Rokh et~al.(2022)Rokh, Azarpeyvand, and
  Khanteymoori]{quantization-survey}
Babak Rokh, Ali Azarpeyvand, and Alireza Khanteymoori.
\newblock A comprehensive survey on model quantization for deep neural
  networks.
\newblock \emph{CoRR}, 2022.

\bibitem[Romero et~al.(2015)Romero, Ballas, Kahou, Chassang, Gatta, and
  Bengio]{fitnet}
Adriana Romero, Nicolas Ballas, Samira~Ebrahimi Kahou, Antoine Chassang, Carlo
  Gatta, and Yoshua Bengio.
\newblock Fitnets: Hints for thin deep nets.
\newblock In \emph{ICLR}, 2015.

\bibitem[Sanh et~al.(2019)Sanh, Debut, Chaumond, and Wolf]{distill-bert}
Victor Sanh, Lysandre Debut, Julien Chaumond, and Thomas Wolf.
\newblock Distilbert, a distilled version of {BERT:} smaller, faster, cheaper
  and lighter.
\newblock \emph{CoRR}, abs/1910.01108, 2019.

\bibitem[Schuermans et~al.(2004)Schuermans, Lemmerling, and
  Huffel]{low-rank-old1}
M. Schuermans, Philippe Lemmerling, and Sabine~Van Huffel.
\newblock Structured weighted low rank approximation.
\newblock \emph{Numerical Linear Algebra with Applications}, 11\penalty0
  (5-6):\penalty0 609--618, 2004.

\bibitem[Srebro and Jaakkola(2003)]{low-rank-old2}
Nathan Srebro and Tommi~S. Jaakkola.
\newblock Weighted low-rank approximations.
\newblock In \emph{ICML}, pages 720--727. {AAAI} Press, 2003.

\bibitem[Strohman and Croft(2007)]{pre-dl2}
Trevor Strohman and W.~Bruce Croft.
\newblock Efficient document retrieval in main memory.
\newblock In \emph{SIGIR}, pages 175--182. {ACM}, 2007.

\bibitem[Sun et~al.(2019)Sun, Cheng, Gan, and Liu]{pkd}
Siqi Sun, Yu Cheng, Zhe Gan, and Jingjing Liu.
\newblock Patient knowledge distillation for {BERT} model compression.
\newblock In \emph{EMNLP}, pages 4322--4331. ACL, 2019.

\bibitem[Sung et~al.(2022)Sung, Cho, and Bansal]{adapter}
Yi{-}Lin Sung, Jaemin Cho, and Mohit Bansal.
\newblock {VL-ADAPTER:} parameter-efficient transfer learning for
  vision-and-language tasks.
\newblock In \emph{CVPR}, pages 5217--5227. {IEEE}, 2022.

\bibitem[Tai et~al.(2016)Tai, Xiao, Wang, and E]{lr-cnn2}
Cheng Tai, Tong Xiao, Xiaogang Wang, and Weinan E.
\newblock Convolutional neural networks with low-rank regularization.
\newblock In \emph{ICLR}, 2016.

\bibitem[Tan and Bansal(2019)]{lxmert}
Hao Tan and Mohit Bansal.
\newblock {LXMERT:} learning cross-modality encoder representations from
  transformers.
\newblock In \emph{EMNLP}, pages 5099--5110. ACL, 2019.

\bibitem[Touvron et~al.(2021{\natexlab{a}})Touvron, Cord, Douze, Massa,
  Sablayrolles, and J{\'{e}}gou]{deit}
Hugo Touvron, Matthieu Cord, Matthijs Douze, Francisco Massa, Alexandre
  Sablayrolles, and Herv{\'{e}} J{\'{e}}gou.
\newblock Training data-efficient image transformers {\&} distillation through
  attention.
\newblock In \emph{ICML}, pages 10347--10357. {PMLR}, 2021{\natexlab{a}}.

\bibitem[Touvron et~al.(2021{\natexlab{b}})Touvron, Cord, El-Nouby, Bojanowski,
  Joulin, Synnaeve, and J{\'e}gou]{patchconv}
Hugo Touvron, Matthieu Cord, Alaaeldin El-Nouby, Piotr Bojanowski, Armand
  Joulin, Gabriel Synnaeve, and Herv{\'e} J{\'e}gou.
\newblock Augmenting convolutional networks with attention-based aggregation.
\newblock \emph{arXiv preprint arXiv:2112.13692}, 2021{\natexlab{b}}.

\bibitem[Touvron et~al.(2022)Touvron, Cord, and J{\'e}gou]{deit3}
Hugo Touvron, Matthieu Cord, and Herv{\'e} J{\'e}gou.
\newblock Deit iii: Revenge of the vit.
\newblock In \emph{ECCV}, pages 516--533. Springer, 2022.

\bibitem[Vaswani et~al.(2017)Vaswani, Shazeer, Parmar, Uszkoreit, Jones, Gomez,
  Kaiser, and Polosukhin]{transformer}
Ashish Vaswani, Noam Shazeer, Niki Parmar, Jakob Uszkoreit, Llion Jones,
  Aidan~N. Gomez, Lukasz Kaiser, and Illia Polosukhin.
\newblock Attention is all you need.
\newblock In \emph{NIPS}, pages 5998--6008, 2017.

\bibitem[Wang et~al.(2022{\natexlab{a}})Wang, Qin, Bai, Zhang, and
  Fu]{prune-survey}
Huan Wang, Can Qin, Yue Bai, Yulun Zhang, and Yun Fu.
\newblock Recent advances on neural network pruning at initialization.
\newblock In \emph{IJCAI}, pages 5638--5645. ijcai.org, 2022{\natexlab{a}}.

\bibitem[Wang et~al.(2022{\natexlab{b}})Wang, Yang, Men, Lin, Bai, Li, Ma,
  Zhou, Zhou, and Yang]{ofa}
Peng Wang, An Yang, Rui Men, Junyang Lin, Shuai Bai, Zhikang Li, Jianxin Ma,
  Chang Zhou, Jingren Zhou, and Hongxia Yang.
\newblock {OFA:} unifying architectures, tasks, and modalities through a simple
  sequence-to-sequence learning framework.
\newblock In \emph{ICML}, pages 23318--23340. {PMLR}, 2022{\natexlab{b}}.

\bibitem[Wang et~al.(2020)Wang, Li, Khabsa, Fang, and Ma]{lin-former}
Sinong Wang, Belinda~Z. Li, Madian Khabsa, Han Fang, and Hao Ma.
\newblock Linformer: Self-attention with linear complexity.
\newblock \emph{CoRR}, abs/2006.04768, 2020.

\bibitem[Wei et~al.(2022)Wei, Tay, Bommasani, Raffel, Zoph, Borgeaud, Yogatama,
  Bosma, Zhou, Metzler, Chi, Hashimoto, Vinyals, Liang, Dean, and
  Fedus]{emergent}
Jason Wei, Yi Tay, Rishi Bommasani, Colin Raffel, Barret Zoph, Sebastian
  Borgeaud, Dani Yogatama, Maarten Bosma, Denny Zhou, Donald Metzler, Ed~H.
  Chi, Tatsunori Hashimoto, Oriol Vinyals, Percy Liang, Jeff Dean, and William
  Fedus.
\newblock Emergent abilities of large language models.
\newblock \emph{CoRR}, 2022.

\bibitem[Weng et~al.(2020)Weng, Zhao, Liu, Chen, Lin, and
  Daniel]{weight-perturb}
Tsui{-}Wei Weng, Pu Zhao, Sijia Liu, Pin{-}Yu Chen, Xue Lin, and Luca Daniel.
\newblock Towards certificated model robustness against weight perturbations.
\newblock In \emph{AAAI}, pages 6356--6363. {AAAI} Press, 2020.

\bibitem[Xiao et~al.(2018)Xiao, Liu, Zhou, Jiang, and Sun]{upernet}
Tete Xiao, Yingcheng Liu, Bolei Zhou, Yuning Jiang, and Jian Sun.
\newblock Unified perceptual parsing for scene understanding.
\newblock In \emph{ECCV}, pages 418--434. Springer, 2018.

\bibitem[Xie et~al.(2017)Xie, Girshick, Doll{\'a}r, Tu, and He]{resnext}
Saining Xie, Ross Girshick, Piotr Doll{\'a}r, Zhuowen Tu, and Kaiming He.
\newblock Aggregated residual transformations for deep neural networks.
\newblock In \emph{CVPR}, pages 1492--1500. IEEE, 2017.

\bibitem[Xu et~al.(2021)Xu, Luo, Zhang, Tan, Chang, Huang, and
  Huang]{prune-unstructured}
Runxin Xu, Fuli Luo, Zhiyuan Zhang, Chuanqi Tan, Baobao Chang, Songfang Huang,
  and Fei Huang.
\newblock Raise a child in large language model: Towards effective and
  generalizable fine-tuning.
\newblock In \emph{EMNLP}, pages 9514--9528. ACL, 2021.

\bibitem[Xu et~al.(2020)Xu, Li, Zhang, Wen, Wang, Qi, Chen, Lin, and
  Xiong]{lr-dynamic2}
Yuhui Xu, Yuxi Li, Shuai Zhang, Wei Wen, Botao Wang, Yingyong Qi, Yiran Chen,
  Weiyao Lin, and Hongkai Xiong.
\newblock {TRP:} trained rank pruning for efficient deep neural networks.
\newblock In \emph{IJCAI}, pages 977--983. ijcai.org, 2020.

\bibitem[Xue et~al.(2023)Xue, Gao, Li, Qiao, Sun, Li, and Luo]{maskalign}
Hongwei Xue, Peng Gao, Hongyang Li, Yu Qiao, Hao Sun, Houqiang Li, and Jiebo
  Luo.
\newblock Stare at what you see: Masked image modeling without reconstruction.
\newblock In \emph{CVPR}, pages 22732--22741. {IEEE}, 2023.

\bibitem[Yang et~al.(2020)Yang, Tang, Wen, Yan, Hu, Li, Li, and
  Chen]{lr-dynamic1}
Huanrui Yang, Minxue Tang, Wei Wen, Feng Yan, Daniel Hu, Ang Li, Hai Li, and
  Yiran Chen.
\newblock Learning low-rank deep neural networks via singular vector
  orthogonality regularization and singular value sparsification.
\newblock In \emph{CVPR Workshops}, pages 2899--2908. {IEEE}, 2020.

\bibitem[Yang et~al.(2022)Yang, Li, Zeng, Li, Yuan, and Li]{kd-abandon}
Zhendong Yang, Zhe Li, Ailing Zeng, Zexian Li, Chun Yuan, and Yu Li.
\newblock Vitkd: Practical guidelines for vit feature knowledge distillation.
\newblock \emph{CoRR}, abs/2209.02432, 2022.

\bibitem[Yu et~al.(2017)Yu, Liu, Wang, and Tao]{lr-dynamic3}
Xiyu Yu, Tongliang Liu, Xinchao Wang, and Dacheng Tao.
\newblock On compressing deep models by low rank and sparse decomposition.
\newblock In \emph{CVPR}, pages 67--76. {IEEE}, 2017.

\bibitem[Yuan et~al.(2021)Yuan, Chen, Wang, Yu, Shi, Jiang, Tay, Feng, and
  Yan]{t2t-vit}
Li Yuan, Yunpeng Chen, Tao Wang, Weihao Yu, Yujun Shi, Zi-Hang Jiang,
  Francis~EH Tay, Jiashi Feng, and Shuicheng Yan.
\newblock Tokens-to-token vit: Training vision transformers from scratch on
  imagenet.
\newblock In \emph{ICCV}, pages 558--567. IEEE, 2021.

\bibitem[Zaken et~al.(2022)Zaken, Goldberg, and Ravfogel]{bias-only}
Elad~Ben Zaken, Yoav Goldberg, and Shauli Ravfogel.
\newblock Bitfit: Simple parameter-efficient fine-tuning for transformer-based
  masked language-models.
\newblock In \emph{ACL}, pages 1--9. ACL, 2022.

\bibitem[Zhang et~al.(2016)Zhang, Zou, He, and Sun]{lr-cnn3}
Xiangyu Zhang, Jianhua Zou, Kaiming He, and Jian Sun.
\newblock Accelerating very deep convolutional networks for classification and
  detection.
\newblock \emph{TPAMI}, 38\penalty0 (10):\penalty0 1943--1955, 2016.

\bibitem[Zhang et~al.(2020)Zhang, Zhao, Lin, Zhu, and He]{ccl}
Zhu Zhang, Zhou Zhao, Zhijie Lin, Jieming Zhu, and Xiuqiang He.
\newblock Counterfactual contrastive learning for weakly-supervised
  vision-language grounding.
\newblock In \emph{NeurIPS}, 2020.

\bibitem[Zhou et~al.(2019)Zhou, Zhao, Puig, Xiao, Fidler, Barriuso, and
  Torralba]{ade20k}
Bolei Zhou, Hang Zhao, Xavier Puig, Tete Xiao, Sanja Fidler, Adela Barriuso,
  and Antonio Torralba.
\newblock Semantic understanding of scenes through the ade20k dataset.
\newblock \emph{IJCV}, 127:\penalty0 302--321, 2019.

\end{thebibliography}
}

\end{document}